\documentclass[runningheads]{llncs}
\usepackage{epstopdf}
\usepackage{tabularx,threeparttable}
\usepackage{array}
\usepackage[hidelinks]{hyperref}
\usepackage{graphicx}
\usepackage{subfigure}
\usepackage{amssymb, amsmath, bm}
\usepackage{mathtools}
\usepackage{mathrsfs}
\usepackage{booktabs}
\usepackage{multirow}
\usepackage{cite}
\usepackage{url}
\usepackage{color}
\usepackage{dsfont}

\def\ie{\emph{i.e.}}

\def\etal{{\em et al.}}

\newcolumntype{?}[1]{!{\vrule width #1}}
\makeatletter
\newcommand\footnoteref[1]{\protected@xdef\@thefnmark{\ref{#1}}\@footnotemark}
\makeatother

\makeatletter
\newcommand{\printfnsymbol}[1]{%
	\textsuperscript{\@fnsymbol{#1}}%
}
\makeatother
\begin{document}
	
\title{Boundary and Entropy-driven Adversarial Learning for Fundus Image Segmentation}
	\author{Shujun Wang\inst{1}\thanks{Equal contribution}\and Lequan Yu\inst{1}\printfnsymbol{1}\and Kang Li\inst{1}\and Xin Yang\inst{1}\and \\Chi-Wing Fu\inst{1,2} \and Pheng-Ann Heng\inst{1,2}
} 
\authorrunning{S. Wang et al.}
	\institute{Department of Computer Science and Engineering, \\The Chinese University of Hong Kong, Hong Kong SAR, China
		\email{sjwang@cse.cuhk.edu.hk}
	\and Guangdong Provincial Key Laboratory of Computer Vision and Virtual Reality Technology, Shenzhen Institutes of Advanced Technology, \\Chinese Academy of Sciences, Shenzhen, China
	}

\maketitle              

\begin{abstract}
	Accurate segmentation of the optic disc (OD) and cup (OC) in fundus images from different datasets is critical for glaucoma disease screening. 
	The cross-domain discrepancy (domain shift) hinders the generalization of deep neural networks to work on different domain datasets. 
	In this work, we present an unsupervised domain adaptation framework, called \textit{Boundary and Entropy-driven Adversarial Learning (BEAL)}, to improve the OD and OC segmentation performance, especially on the ambiguous boundary regions.
	In particular, our proposed BEAL framework utilizes the adversarial learning to encourage the boundary prediction and mask probability entropy map (uncertainty map) of the target domain to be similar to the source ones,  generating more accurate boundaries and suppressing the high uncertainty predictions of OD and OC segmentation.
	We evaluate the proposed BEAL framework on two public retinal fundus image datasets (Drishti-GS and RIM-ONE-r3), and the experiment results demonstrate that our method outperforms the state-of-the-art unsupervised domain adaptation methods. Our code is available at \url{https://github.com/EmmaW8/BEAL}.

	\keywords{Unsupervised domain adaptation \and Optic disc and cup segmentation \and Fundus images \and Adversarial learning}
\end{abstract}

\section{Introduction}
Automated segmentation of the optic disc (OD) and cup (OC) from fundus images is beneficial to glaucoma screening and diagnosis~\cite{fu2018joint}.
Deep convolutional neural networks (CNNs) have brought significant improvement for automated OD and OC segmentation under the supervised learning setting, but fail to generate satisfactory predictions on new datasets due to cross-domain discrepancy (domain shift)~\cite{kamnitsas2017unsupervised}.
For instance, the M-Net~\cite{fu2018joint} achieves the state-of-the-art performance on the ORIGA testing dataset, but has poor generalization ability to work on other testing datasets~\cite{wang2019patch}.

\begin{figure*}[!t]
	\centering
	\includegraphics[width=0.95\linewidth]{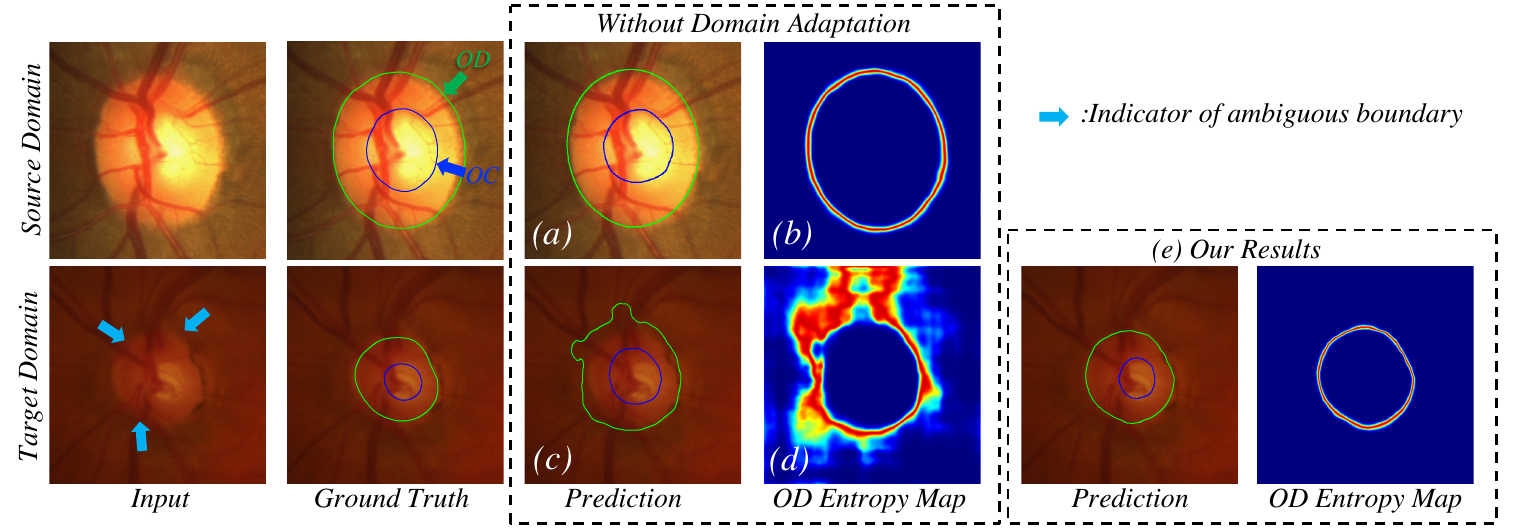}
	\caption{
		Comparison of the OD and OC predictions and the entropy maps of OD.
		The middle two columns show results on source and target domain images of the model trained without domain adaptation.
		The right most two columns show the results of our method on the same target domain image.				
		Red color in the entropy maps ((b) and (d)) indicates high entropy values. 
	}
	\label{Fig:background}
	\centering
\end{figure*}

Very recently, unsupervised domain adaptation methods have been explored to deal with the performance degradation caused by the domain shift in medical imaging community, since acquiring extra annotations on the target domain is time- and money-consuming. 
Some of the previous unsupervised domain adaptation methods improved the performance of network on a target domain by transferring the input images from the target domain to the source domain, and then applying the network trained on the source domain to transferred images~\cite{zhang2018task,chen2018semantic}.
Without any paired images, Cycle-GAN~\cite{zhu2017unpaired} and its variants were the popular methods to transfer image appearance.
Besides, high-level feature alignment was used to explore the shared hidden feature space between different domain datasets and aimed to generate similar predictions for both datasets~\cite{kamnitsas2017unsupervised,dou2018unsupervised}.
Recently, output space alignment was exploited to incorporate the spatial and geometry structures information of predictions~\cite{tsai2018learning,wang2019patch}.
For example, Wang~\etal~\cite{wang2019patch} presented a novel patch-based output space adversarial learning framework to jointly segment the OD and OC from different fundus image datasets.
However, most previous methods fail to produce reliable predictions on soft boundary regions of the target domain images, \ie, the areas among different structures without clear boundary, due to the large appearance difference between the source and target domain images and the low intensity contrast between different structures. 
Therefore, developing an effective domain adaptation method to improve the prediction performance on soft boundary regions of the target domain images is still a challenging problem.

In this work, we present a novel unsupervised domain adaptation framework, called \textit{Boundary and Entropy-driven Adversarial Learning (BEAL)}, to improve the accuracy to segment the OD and OC over different fundus image datasets.
Our method is based on two main observations.
First, deep networks trained on the source domain tend to generate ambiguous and inaccurate boundaries for target domain images, while the boundary prediction of source domain is more structured (\ie,~relative position and shape); see Figs.~\ref{Fig:background}(a) and (c).
Therefore, an effective way to improve the accuracy of target domain predictions is to perform a boundary-driven adversarial learning, which enforces domain-invariant boundary structure between the source and target domains.
Second, the network is prone to generate certainty (low-entropy) predictions on the source domain images\cite{vu2018advent}, resulting in a clear prediction entropy map with high entropy values only along the object boundaries, as shown in Fig.~\ref{Fig:background}(b).
While the predictions of target domain are uncertain, and the entropy map of mask prediction is noisy with high entropy outputs; see the OD entropy map in Fig.~\ref{Fig:background}(d).
Accordingly, enforcing certainty predictions (low-entropy) on the target domain becomes a feasible solution to improve the target domain segmentation performance.
Based on these observations, we develop a boundary and entropy-driven adversarial learning method to segment the OD and OC from the target domain fundus images by generating accurate boundaries and suppressing the high uncertainty regions; see our results in Fig.~\ref{Fig:background}(e).
Specifically, we exploit the adversarial learning technique to simultaneously encourage the boundary and entropy map predictions to be domain-invariant simultaneously.
The proposed method was extensively evaluated on two public fundus image datasets, \ie, RIM-ONE-r3~\cite{fumero2011rim} and Drishti-GS~\cite{sivaswamy2015comprehensive}, demonstrating state-of-the-art results.
We also conducted an ablation study to show the effectiveness of each component in our method.

\section{Methodology}
\begin{figure*}[!t]
	\centering
	\includegraphics[width=0.95\linewidth]{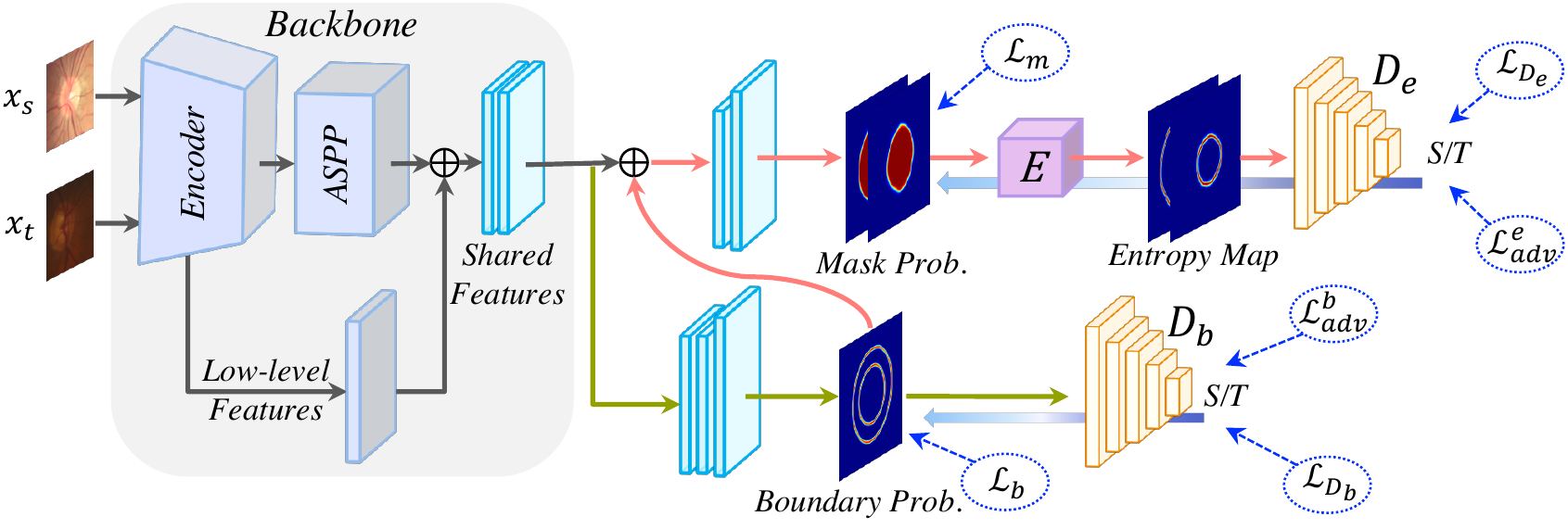}
	\caption{Overview of our BEAL framework for unsupervised domain adaptation. 
		The backbone is based on the DeepLabv3+ \cite{chen2018encoder} architecture with Atrous Spatial Pyramid Pooling (ASPP) component followed by boundary and mask branches. 
		We then apply Shannon Entropy ($E$) to obtain the entropy maps. 
		Finally, we add two discriminators to apply adversarial learning on the boundary and entropy maps. 
	}
	\label{fig:framework}
	\centering
\end{figure*}

Fig.~\ref{fig:framework} overviews our proposed BEAL framework for segmenting OD and OC in fundus images from different domains
to segment OD and OC in fundus images from different domains. 
The key technical contribution in our method is a boundary and entropy-driven adversarial learning framework for accurate and confident predictions on the target domain.

\subsection{Boundary-driven Adversarial Learning (BAL)}
For target domain images, the segmentation network optimized by source domain supervision tends to generate ambiguous and unstructured predictions.
To mitigate this problem, we formulate a boundary-driven adversarial learning model to enforce the predicted boundary structure in the target domain to be similar to that in the source domain. 
Specifically, we adopt a boundary prediction branch to regress the boundary and a mask prediction branch for the OD and OC segmentation by changing the decoder of the segmentation network (Network details will be presented later).
Then, we introduce an adversarial learning model by taking the regressed boundary as input.

Formally, we consider a source domain image set $\mathcal{I}_{S} \subset \mathbb{R}^{H\times W\times 3}$ along with ground truth segmentation maps $\mathcal{Y}_{S} \subset \mathbb{R}^{H\times W}$, and another target domain image set $\mathcal{I}_{T} \subset \mathbb{R}^{H\times W\times 3}$ without any ground truth.
For each source domain input image $x_{s} \in \mathcal{I}_{S}$, our network produces the boundary prediction $p^{b}_{x_{s}}$ and mask probability prediction $p^m_{x_{s}}$.
Similarly, our network also generates the boundary prediction $p^{b}_{x_{t}}$ and mask prediction $p^m_{x_{t}}$ for each target domain input image $x_{t} \in \mathcal{I}_{T}$.
To use the boundary to drive the adversarial learning model, we utilize a boundary discriminator $D_b$ to align the distributions of the boundary predictions ($p^{b}_{x_s}$, $p^{b}_{x_t}$).
The discriminator network $D_{b}$ aims to figure out whether the boundary is from the source or from the target domain.
So, the training objective for the boundary discriminator is formulated as 
\begin{equation}
\label{equ:dis_boundary}
\mathcal{L}_{D_b} = \frac{1}{N}\sum_{x_s \in \mathcal{I}_{S}}\mathcal{L}_{D}(p^{b}_{x_s}, 1) +\frac{1}{M} \sum_{x_t \in \mathcal{I}_{T}}\mathcal{L}_{D}(p^{b}_{x_t}, 0),
\end{equation}
where $\mathcal{L}_{D}$ is the binary cross-entropy loss, and $N$ and $M$ are the total number of source and target domain images, respectively. 
To further align the boundary structure distribution, we utilize the adversarial learning to optimize the segmentation network with the boundary adversarial objective   
\begin{equation}
\label{equ:adv_boundary}
\mathcal{L}_{adv}^{b} = \frac{1}{M}\sum_{x_t \in \mathcal{I}_{T}} \mathcal{L}_{D}(p^{b}_{x_t}, 1).
\end{equation}

\subsection{Entropy-driven Adversarial Learning (EAL)}
With the boundary-driven adversarial learning model, the predictions on the target domain are still prone to be \textit{high-entropy} (under-confident) on the soft boundary regions.
To suppress uncertain predictions, we further adopt an entropy-driven adversarial learning model to narrow down the performance gap between the source and target domains by enforcing the entropy maps of the target domain predictions to be similar to the source ones.
In detail, given the pixel-wise mask probability prediction $p^{m}_x$ of input image $x$, we use the Shannon Entropy to calculate the entropy map in pixel level \cite{vu2018advent} following
\begin{equation}
\label{equ:entropy}
E(x)=p^{m}_{x} \cdot \text{log}(p^{m}_{x}).
\end{equation}

To conduct the entropy-driven adversarial learning, we construct an entropy discriminator network $D_e$ to align the distributions of entropy maps $E(x_s)$ and $E(x_t)$.
Similar to boundary-driven adversarial learning, we train the entropy discriminator to figure out whether the entropy map is from the source or the target domain.
Specifically, the objective function of $D_e$ is
\begin{equation}
\label{equ:dis_entropy}
\mathcal{L}_{D_e} = \frac{1}{N}\sum_{x_s\in \mathcal{I}_{S}}\mathcal{L}_{D}(E(x_s), 1) + \frac{1}{M}\sum_{x_t \in \mathcal{I}_{T}}\mathcal{L}_{D}(E(x_t), 0).
\end{equation}
At the same time, we optimize the segmentation network to fool the discriminator using the following adversarial loss
\begin{equation}
\label{equ:adv_entropy}
\mathcal{L}_{adv}^{e} = \frac{1}{M}\sum_{x_t \in \mathcal{I}_{T}} \mathcal{L}_{D}(E(x_t), 1),
\end{equation}
which encourages the segmentation network to generate prediction entropy on the target domain images similar to the source domain ones.

\subsection{Network Architecture and Training Procedure}
We use an adapted DeepLabv3+ \cite{chen2018encoder} as the segmentation backbone of our BEAL framework.
Specifically, we replace the Xception with a lightweight and handy MobileNetV2 to reduce the number of parameters and accelerate the computation,
and add the boundary and mask prediction branches after the high-level and low-level feature concatenation.
The boundary branch consists of three convolutional layers with output channel of \{256, 256, 1\} followed by ReLU and batch normalization, except the last one with Sigmoid activation.
The mask branch has one convolutional layer with taking the concatenation of boundary predictions and shared features as input.
The final prediction are obtained after bilinear interpolation to the same size of the input image.
The discriminators consist of five convolutional layers following the previous work~\cite{wang2019patch}.

We optimize the segmentation network and the discriminators in an alternate way.
To optimize the boundary and entropy discriminators, we minimize the objective function in Eqs.~\eqref{equ:dis_boundary} and \eqref{equ:dis_entropy}, respectively.
To optimize the segment network, we calculate the mask prediction loss $\mathcal{L}_{m}$ and the boundary regression loss $\mathcal{L}_b$ on the source domain images, and the adversarial loss $\mathcal{L}_{adv}^b$ and $\mathcal{L}_{adv}^e$ on the target domain images. The overall objective of segmentation network is
\begin{align}
\label{equ:seg}
\mathcal{L} = &\mathcal{L}_{m} + \mathcal{L}_{b} + \lambda(\mathcal{L}_{adv}^b + \mathcal{L}_{adv}^e), \\
\mathcal{L}_{b} &=   \frac{1}{N} \sum _{x_s \in \mathcal{I_S}} (y_{x_s}^{b}-p_{x_s}^{b})^2, \nonumber \\
\text{and}  \ \mathcal{L}_{m} = - \frac{1}{N} \sum _{x_s \in \mathcal{I_S}} & [y_{x_s}^{m} \cdot log(p_{x_s}^{m})  + (1-y_{x_s}^{m})\cdot log(1-p_{x_s}^{m})], \nonumber
\end{align}
where $y^{m}$ and $y^{b}$ are the ground truth of the mask and boundary, respectively, and $\lambda$ is a balance coefficient.
We formulate the mask prediction as a multi-label learning~\cite{wang2019patch} and generate the probability maps of OD and OC simultaneously. 
We take the entropy map of OD and OC together as the discriminator input.
To acquire the boundary ground truth, we apply the Sobel operation and Gaussian filter to the ground truth masks.

\section{Experiments and Results}

\subsubsection{Dataset.}
To evaluate our method, we utilize the training part of the REFUGE challenge dataset\footnote{\label{note1}\url{https://refuge.grand-challenge.org/}} as the source domain, and the public Drishti-GS \cite{sivaswamy2015comprehensive}, and the RIM-ONE-r3 \cite{fumero2011rim} dataset as the target domains including both the training and testing parts.
The detailed statistics of the datasets are shown in Table~\ref{tab:datasetstatistics}.

\begin{table*} [!tbp]
	\centering
	\caption{Statistics of the datasets used in evaluating our method.}
	\label{tab:datasetstatistics}
	\begin{tabular}{p{1.8cm}<{\centering}|p{5cm}<{\centering}|p{3.5cm}<{\centering}}
		\toprule[1pt]
		Domain & Dataset &  Number of samples   \\
		\hline
		Source & REFUGE\footnoteref{note1} (Train) &  400 \\
		\hline
		Target & RIM-ONE-r3\cite{fumero2011rim} (Train + Test) &99 + 60\\
		\hline
		Target & Drishti-GS\cite{sivaswamy2015comprehensive} (Train + Test) &  50 + 51  \\
		\toprule[1pt]
	\end{tabular}
\end{table*}

\begin{table*} [!tbp]
	\centering
	\caption{Comparison with other methods on the target domain datasets.
	}
	\label{table:results}
	\begin{tabular}{p{3.5cm}<{\centering}|p{2cm}<{\centering}|p{2cm}<{\centering}|p{2cm}<{\centering}|p{2cm}<{\centering}}
		\toprule[1pt]
		\multirow{2}{*}{\textbf{Method}} & \multicolumn{2}{c|}{\textbf{RIM-ONE-r3}\cite{fumero2011rim}} & \multicolumn{2}{c}{\textbf{Drishti-GS}\cite{sivaswamy2015comprehensive}} \\
		\cline{2-5} & $DI_{cup}$  & $DI_{disc}$   &  $DI_{cup}$  & $DI_{disc}$   \\
		\hline
		w/o DA&  {0.744} &{0.779}   &  {0.836} & {0.944}  \\
		\hline
		Upper bound & 0.856 &0.968&0.901&0.974\\\hline \hline
		TD-GAN\cite{zhang2018task} &0.728&0.853&  0.747 & 0.924  \\ \hline
		Hoffman~\etal~\cite{hoffman2016fcns}&0.755&0.852  &0.851&0.959  \\ \hline
		Javanmardi~\etal~\cite{javanmardi2018domain} &0.779&0.853 &  0.849 & 0.961  \\ \hline
		OSAL-pixel~\cite{wang2019patch}   & 0.778&0.854   &  0.851 & 0.962  \\  \hline
		\textit{p}OSAL~\cite{wang2019patch}  &  {0.787} &{0.865} &  {0.858} & \textbf{0.965}  \\ \hline
		\textbf{BEAL} (ours)&\textbf{0.810}&\textbf{0.898}  &\textbf{0.862}&0.961\\ \hline 
		\toprule[1pt]
	\end{tabular}
\end{table*}

\subsubsection{Implementation details.}
Our framework was implemented with the PyTorch library.
We trained the whole framework directly without the warm-up phase of supervised learning in a minibatch of size 8. 
The discriminator $D_{e}$ and $D_b$ were optimized with the SGD algorithm, while the Adam optimizer was utilized for optimizing the segmentation network.
We set the initial learning rate of SGD as $1e-3$ and divided it by 0.2 every 100 epochs for a total of 200 epochs.
The learning rate of discriminator training was set as $2.5e-5$.
We cropped $512 \times 512$ ROIs centering at OD as the network input following the previous work \cite{wang2019patch} by utilizing a simple U-Net architecture. We used the standard data augmentation, including random rotation, flipping, elastic transformation, contrast adjustment, adding Gaussian noise, and random erasing~\cite{wang2019patch}.

\subsubsection{Quantitative analysis.}
We use the dice coefficients ($DI$) of OD and OC to quantitatively evaluate the results produced from our method.
The segmentation results of our approach and others on RIM-ONE-r3 and Drishti-GS are presented in Table~\ref{table:results}.
We compare our framework with the baseline (w/o DA), the supervised method (\textit{Upper bound}), and other unsupervised domain adaptation methods, including TD-GAN \cite{zhang2018task}, high-level feature alignment~\cite{hoffman2016fcns} and output space-based adaptation~\cite{javanmardi2018domain,wang2019patch}.
The results of other methods are inherited from the previous work~\cite{wang2019patch}. 
Compared with the state-of-the-art unsupervised domain adaptation method \textit{p}OSAL, our BEAL framework achieves $2.3\%$ and $3.3\%$ DI improvement for the OC and OD segmentation on the RIM-ONE-r3 dataset, demonstrating the effectiveness of the boundary and entropy-driven domain adaption method.
Since the domain distribution gap between the REFUGE and Drishti-GS data is smaller than the difference between REFUGE and RIM-ONE-r3 data~\cite{wang2019patch}, the absolute DI values of optic cup and disc on Drishti-GS is higher than that on RIM-ONE-r3.
Therefore, the room for improvement on the Drishti-GS dataset is limited, as the current performance is approaching the upper bound.
Nevertheless, our method still outperforms the state-of-the-arts for the cup segmentation, and achieves comparable results with \textit{p}OSAL for the disc segmentation on the Drishti-GS dataset, demonstrating the effectiveness of our method to handle with varying degrees of domain shifts.

\subsubsection{Qualitative analysis.}
\begin{figure*}[!t]
	\centering
	\includegraphics[width=0.99\linewidth]{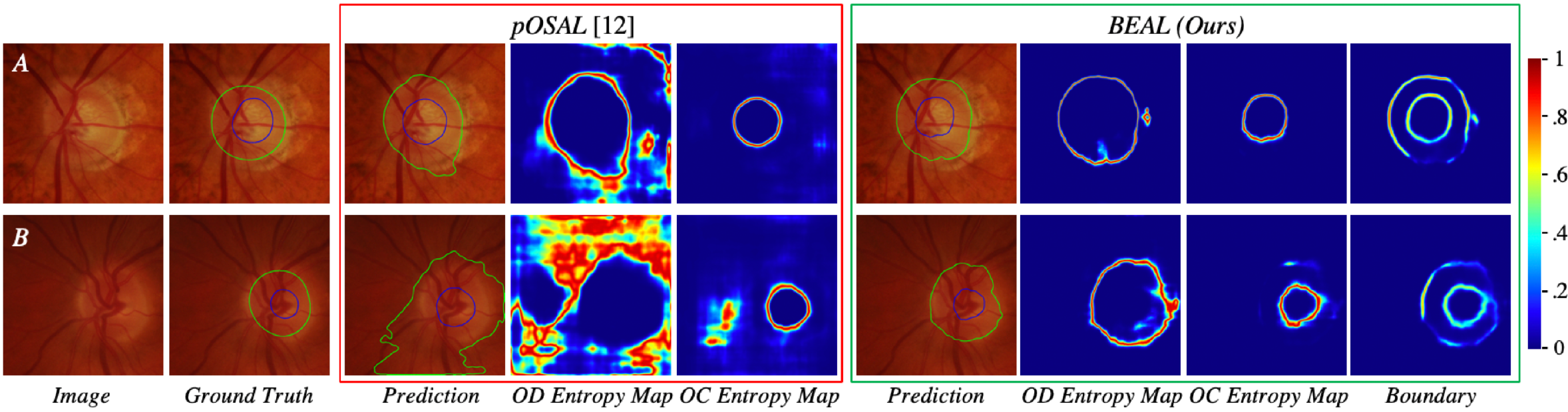}
	\caption{Qualitative results of \textit{p}OSAL \cite{wang2019patch} and our method on the RIM-ONE-r3 dataset \cite{fumero2011rim}. 
		Our method can improve the segmentation results with accurate boundary, and generate clear prediction entropy maps. Green and blue lines represent the disc and cup contours, respectively. The entropy values are rescaled to [0,1] for better visualization.
	}
	\label{fig:results}
	\centering
\end{figure*}

We show some visual results of the OD and OC segmentation, prediction entropy map, and predicted boundary on the RIM-ONE-r3 dataset in Fig.~\ref{fig:results}.
It shows that the \textit{p}OSAL hardly predicts accurate boundary on the ambiguous regions and generates high entropy values. 
By leveraging the proposed boundary and entropy-driven adversarial learning, our method produces more accurate boundaries and clean entropy maps of the mask predictions.

\begin{table*} [!htbp]
	\centering
	\caption{Ablation study on different components.}
	\label{table:ablationstudy}
	\begin{tabular}{p{4cm}<{\centering}|p{1.8cm}<{\centering}|p{1.8cm}<{\centering}|p{1.8cm}<{\centering}|p{1.8cm}<{\centering}}
		\toprule[1pt]
		\multirow{2}{*}{\textbf{Method}} & \multicolumn{2}{c|}{\textbf{RIM-ONE-r3}\cite{fumero2011rim}} & \multicolumn{2}{c}{\textbf{Drishti-GS}\cite{sivaswamy2015comprehensive}} \\
		\cline{2-5} & $DI_{cup}$  & $DI_{disc}$  &  $DI_{cup}$  & $DI_{disc}$   \\
		\hline
		Baseline w/o boundary   &  {0.744} &{0.779}  &  {0.836} & {0.944}   \\ \hline
		Baseline  &  {0.779} &{0.885} &  {0.841} & {0.951}    \\  \hline
		Baseline+BAL  &  {0.781} &{0.893}  &  {0.847} & {0.958}  \\ \hline 
		Baseline+EAL  &  {0.800} &\textbf{0.898}  &  {0.851} & {0.960}  \\ \hline
		\textbf{BEAL (ours)}&\textbf{0.810}&\textbf{0.898} &\textbf{0.862}&\textbf{0.961} \\
		\toprule[1pt]
	\end{tabular}
\end{table*}

\subsubsection{Ablation study.}
We conducted a set of ablation experiments to evaluate the effectiveness of each component: 
(i) DeepLabv3+ network  (Baseline w/o boundary),
(ii) DeepLabv3+ network equipped with a boundary branch (Baseline),
(iii) boundary-driven adversarial learning (Baseline+BAL),
(iv) entropy-driven adversarial learning (Baseline+EAL); and
(v) our proposed method (BEAL).
The results are shown in Table~\ref{table:ablationstudy}.
%
With extra constraint information from the boundary prediction, the {Baseline} improves performance for both two datasets compared with Baseline w/o boundary.
With additional adversarial learning model, the results show that both {BAL} and {EAL} improve the OD and OC segmentations on the two datasets.
By combining the two adversarial learning methods, we observe a further improvement in the performance, confirming that the effectiveness of our combined adversarial learning model.

\section{Conclusion}
We proposed a novel boundary and entropy-driven adversarial learning method for the OC and OD segmentation in fundus images from different domains.
To address the domain shift challenge, our method encourages the boundary and the entropy map of prediction simultaneously to be domain-invariant, generating more accurate boundaries and suppressing uncertain predictions of OD and OC.
Our method outperforms the state-of-the-art methods, as clearly demonstrated on the two public fundus segmentation datasets.
It is effective and could be generalized to other unsupervised domain adaptation problems.
\\
\\
\textbf{Acknowledgments.}
The work described in this paper was supported by 973 Program under Project No. 2015CB351706, and Research Grants Council of Hong Kong Special Administrative Region under Project No. CUHK14225616, and Hong Kong Innovation and Technology Fund under Project No. ITS/426/17FP, and National Natural Science Foundation of China under Project No. U1613219.

\bibliographystyle{splncs04}
\bibliography{ref}
\end{document}